\documentclass[runningheads]{llncs}
\usepackage{graphicx}
\usepackage{amsmath,amssymb} % define this before the line numbering.
\usepackage{color}
\usepackage{algorithm}
\usepackage{algorithmic}

\begin{document}
\title{Attribute-Guided Face Generation Using Conditional CycleGAN} 
% Replace with your title

\titlerunning{Attribute-Guided Face Generation Using Conditional CycleGAN}
% Replace with a meaningful short version of your title
%

% \author{Yongyi Lu\inst{1}\orcidID{0000-0003-1398-9965} \and
% Yu-Wing Tai\inst{2}\orcidID{0000-0002-3148-0380} \and
% Chi-Keung Tang\inst{1}\orcidID{0000-0001-6495-3685}}

\author{Yongyi Lu\inst{1} \and
	Yu-Wing Tai\inst{2} \and
	Chi-Keung Tang\inst{1}}

%
%Please write out author names in full in the paper, i.e. full given and family names. 
%If any authors have names that can be parsed into FirstName LastName in multiple ways, please include the correct parsing, in a comment to the volume editors:
%\index{Lastnames, Firstnames}
%(Do not uncomment it, because you may introduce extra index items if you do that, we will use scripts for introducing index entries...)
\authorrunning{Y. Lu, Y. W. Tai and C. K. Tang}
% Replace with shorter version of the author list. If there are more authors than fits a line, please use A. Author et al.
%

\institute{The Hong Kong University of Science and Technology \and
Tencent Youtu\\
\email{\{yluaw,cktang\}@cse.ust.hk, yuwingtai@tencent.com}}
\maketitle              % typeset the header of the contribution
\begin{abstract}
We are interested in attribute-guided face generation: given a
low-res face input image, an attribute vector that can be
extracted from a high-res image (attribute image), our new method generates
a high-res face image for the low-res input that satisfies the
given attributes. To address this problem, we condition the
CycleGAN and propose conditional CycleGAN,
which is designed to 1) handle unpaired training data because
the training low/high-res and high-res attribute images may not
necessarily align with each other, and to 2) allow easy control of the
appearance of the generated face via the input attributes.
We demonstrate high-quality results on the {\em attribute-guided 
	conditional CycleGAN}, which can synthesize realistic face
images with appearance easily controlled by user-supplied attributes
(e.g., gender, makeup, hair color, eyeglasses).  Using the
attribute image as identity to produce the corresponding conditional
vector and by incorporating a face verification network,
the attribute-guided network becomes the {\em identity-guided 
	conditional CycleGAN} which produces high-quality and interesting
results on identity transfer. We demonstrate three applications on identity-guided
conditional CycleGAN: identity-preserving face superresolution, face swapping,
and frontal face generation, which consistently show the advantage of our new method.

\let\thefootnote\relax\footnotetext{* This work was partially done when Yongyi Lu was an intern at Tencent Youtu.}

\end{abstract}
\section{Introduction}

\begin{figure}[htb]
	\begin{center}
		\includegraphics[width=0.7\linewidth]{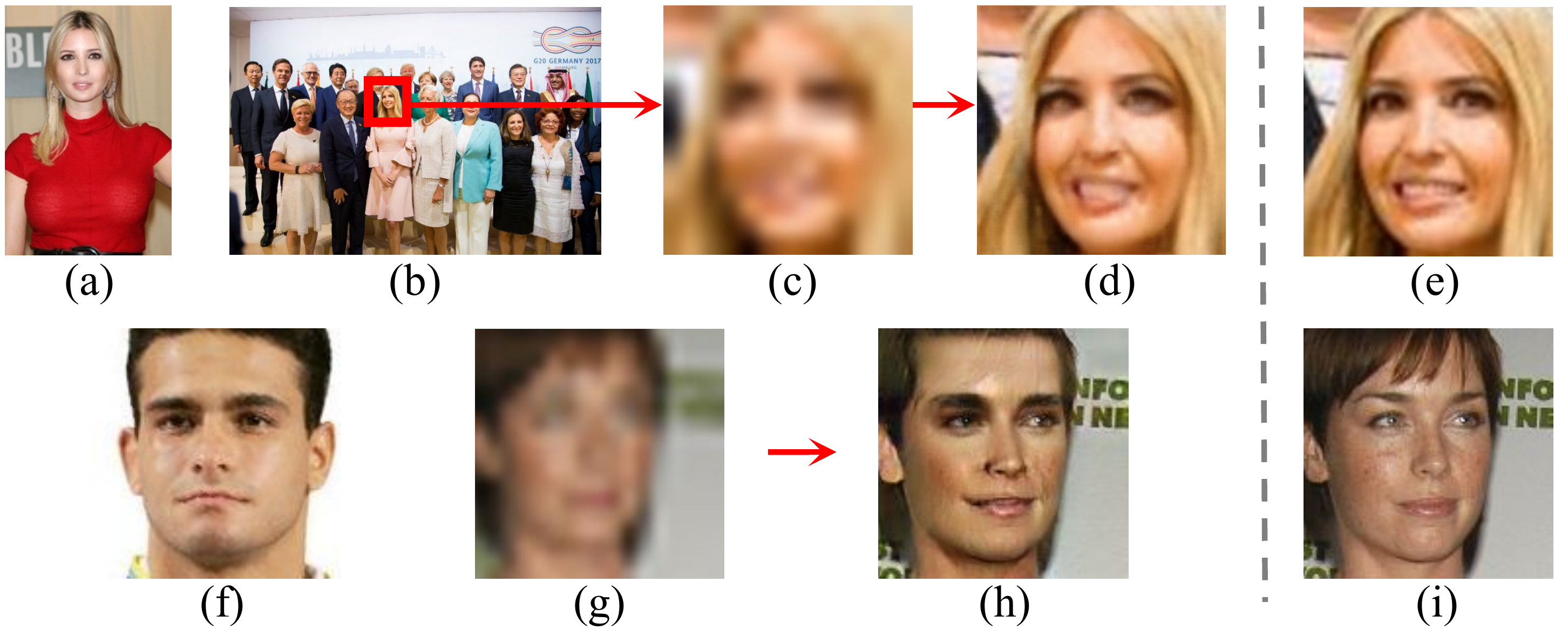} \\
	\end{center}
	\caption{Identity-guided face generation. Top: identity-preserving face super-resolution where (a) is the identity image; (b) input photo; (c) image crop from (b) in low resolution; (d) our generated high-res result; (e) ground truth image. Bottom: face transfer, where (f) is the identity image; (g) input low-res image of another person provides overall shape constraint; (h) our generated high-res result where the man's identity is transferred. To produce the low-res input (g) we down-sample from (i), which is a woman's face.}
	\label{fig:teaser}
\end{figure}

This paper proposes a practical approach, {\em attribute-guided face generation}, 
for natural face image generation where facial appearance can be easily controlled 
by user-supplied attributes.  Figure~\ref{fig:teaser} shows that by simply providing 
a high-res image of Ivanka Trump, our face superresolution result preserves her identity 
which is not necessarily guaranteed by conventional face superresolution (Figure~\ref{fig:teaser}: top row). When the input attribute/identity image is a different person, our method transfers
the man's identity to the high-res result, where the low-res input is originally
downsampled from a woman's face (Figure~\ref{fig:teaser}: bottom row). 

We propose to address our face generation problem using conditional CycleGAN.  The
original unconditional CycleGAN~\cite{CycleGAN2017}, where enforcing
cycle consistency has demonstrated state-of-the-art results in
photographic image synthesis, was designed to handle unpaired training
data.  Relaxing the requirement of paired training data is particularly suitable in our case because the
training low/high-res and high-res attribute images do not need
to align with each other. By enforcing cycle consistency, we are able to learn a bijective mapping, or one-to-one correspondence with unpaired data from the same/different domains.
By simply altering the attribute condition, our approach can be
directly applied to generate high-quality face images that simultaneously
preserve the constraints given in the low-res input while
transferring facial features (e.g.,  gender, hair color, emotion,
sun-glasses) prescribed by input face attributes.

Founded on CycleGAN, we present significant results on both attribute-guided and identity-guided face generation, which we believe is important and timely.
Technically, our contribution consists of the new conditional CycleGAN to
guide the single-image super-resolution process via the embedding of complex
attributes for generating images with high level of photo-realism:

First, in our {\em attribute-guided conditional CycleGAN}, the adversarial
loss is modified to include a conditional
feature vector as part of the input to the generator and intra layer to the
discriminator as well.   Using the trained network we demonstrate impressive
results including gender change, transfer of hair color and facial emotion.

Second, in our {\em identity-guided conditional CycleGAN}, we incorporate
a face verification network to produce the conditional vector, and define
the proposed identity loss in an auxiliary discriminator for
preserving facial identity.  Using the trained network, we demonstrate
realistic results on identity transfer which are robust to pose variations and
partial occlusion.  We demonstrate three applications of identity-guided 
conditional CycleGAN: identity-preserving face superresolution, face swapping,
and frontal face generation.

%-------------------------------------------------------------------------
\section{Related Work}

Recent state-of-the-art image generation techniques have leveraged the 
deep convolutional neural networks (CNNs).   For example, in single-image superresolution (SISR), 
a deep recursive CNN for SISR was proposed in~\cite{DBLP:conf/cvpr/KimLL16}. 
Learning upscaling filters have improved accuracy and
speed~\cite{DBLP:journals/corr/DongLT16,espcn,DBLP:journals/corr/WangWWL16}.
A deep CNN approach was proposed in~\cite{Dong2014_6} using bicubic interpolation. 
The ESPCN~\cite{espcn} performs SR by replacing the deconvolution layer in lieu of 
upscaling layer.  However, many existing CNN-based networks still
generate blurry images.
The SRGAN~\cite{DBLP:journals/corr/LedigTHCATTWS16_7} 
uses the Euclidean distance between the feature maps extracted
from the VGGNet to replace the MSE loss which cannot preserve texture details. 
The SRGAN has improved the perceptual quality of generated SR images.
A deep residual network (ResNet) was proposed in~\cite{DBLP:journals/corr/LedigTHCATTWS16_7}
that produces good results for upscaling factors up to $4$.
In~\cite{DBLP:journals/corr/JohnsonAL16} both the perceptual/feature loss and 
pixel loss are used in training SISR. 

Existing GANs~\cite{GAN,Choi_2018_CVPR,zhang2017stackgan} have generated state-of-the-art results for automatic image generation.  The key of
their success lies in the adversarial loss which forces the generated images 
to be indistinguishable from real images. This is achieved by two competing neural networks, 
the generator and the discriminator.  In particular, the DCGAN~\cite{DBLP:journals/corr/RadfordMC15} 
incorporates deep convolutional neural networks into GANs, and has generated
some of the most impressive realistic images to date.  GANs are however 
notoriously difficult to train: GANs are formulated as a minimax 
``game'' between two networks. In practice, it is hard to keep the generator 
and discriminator in balance, where the optimization can oscillate between 
solutions which may easily cause the generator to collapse.  Among different 
techniques, the conditional GAN~\cite{DBLP:journals/corr/IsolaZZE16} 
addresses this problem by enforcing forward-backward consistency, which 
has emerged to be one of the most effective ways to train GAN.

Forward-backward consistency has been enforced in computer vision algorithms such 
as image registration, shape matching, co-segmentation, to name a few.  In 
the realm of image generation using deep learning, 
using unpaired training data, the CycleGAN~\cite{CycleGAN2017} was proposed to learn 
image-to-image translation from a source domain $X$ to 
a target domain $Y$. In addition to the standard GAN loss respectively 
for $X$ and $Y$, a pair of cycle consistency losses (forward and backward) 
was formulated using L1 reconstruction loss. Similar ideas can also be found in
~\cite{kim2017learning,yi2017dualgan}.
For forward cycle consistency, given $x \in X$ the image 
translation cycle should reproduce $x$. Backward cycle consistency is similar.  In this paper, we propose
conditional CycleGAN for face image generation so that the image generation 
process can preserve (or transfer) facial identity, where the results can
be controlled by various input attributes.  Preserving
facial identity has also been explored in synthesizing the corresponding
frontal face image from a single side-view face image~\cite{2017arXiv170404086H}, where 
the identity preserving loss was defined based on the activations of
the last two layers of the Light CNN~\cite{DBLP:journals/corr/WuHS15_lightcnn}.
In multi-view image generation from a single view~\cite{multiview}, a 
condition image (e.g. frontal view) was used to constrain 
the generated multiple views in their coarse-to-fine framework.
However, facial identity was not explicitly preserved in their results 
and thus many of the generated faces look smeared, although as the 
first generated results of multiple views from single
images, the pertinent results already look quite impressive. 

While our conditional CycleGAN is an image-to-image translation framework,~\cite{Perarnau2016_icgan} factorizes an input image into a latent representation 
$z$ and conditional information $y$ using their respective trained 
encoders.  By changing $y$ into $y'$, the generator network then combines 
the same $z$ and new $y'$ to generate an image that satisfies the 
new constraints encoded in $y'$.  We are inspired by their best 
conditional positioning, that is, where $y'$ should be concatenated 
among all of the convolutional layers.  For SISR, in addition, 
$z$ should represent the embedding for a (unconstrained) high-res image, where the generator can combine with the identity 
feature $y$ to generate the super-resolved result.
In~\cite{2017arXiv170501088L}
the authors proposed to learn the dense correspondence between a pair of input
source and reference, so that visual attributes can be swapped or transferred
between them.   In our identity-guided conditional CycleGAN, the input reference is encoded as a conditional identity feature so that the input source can be transformed to target identity even though they do not have perceptually similar structure. 

\section{Conditional CycleGAN}

\subsection{CycleGAN}

A Generative Adversarial Network~\cite{GAN} (GAN) consists of two neural networks,
a generator $G_{X \rightarrow Y}$ and a discriminator $D_{Y}$, which are iteratively trained in
a two-player minimax game manner. The adversarial loss ${\mathcal{L}}(G_{X \rightarrow Y}, D_{Y})$ is defined as
\begin{equation}
\begin{split}
{\mathcal{L}}(G_{X \rightarrow Y}, D_{Y}) = 
& \min_{\Theta_g} \max_{\Theta_d} \big\{\mathbb{E}_{y} [\log D_{Y}(y)]  \\
& + \mathbb{E}_{x}[\log (1-D_{Y}(G_{X \rightarrow Y}(x)))]\big\}
\end{split}
\label{eq:gan}
\end{equation}
where
$\Theta_g$ and $\Theta_d$ are respectively the parameters of the generator $G_{X \rightarrow Y}$ and discriminator $D_{Y}$, and $x \in X$ and $y \in Y$ denotes the {\em unpaired} training data in source and target domain respectively. ${\mathcal{L}}(G_{Y \rightarrow X}, D_{X})$ is analogously defined. 

In CycleGAN, $X$ and $Y$ are two different image representations, and the CycleGAN learns the translation $X \rightarrow Y$ and $Y \rightarrow X$ simultaneously. Different from ``pix2pix''~\cite{DBLP:journals/corr/IsolaZZE16}, training data in CycleGAN is unpaired. Thus, they introduce Cycle Consistency to enforce forward-backward consistency which can be considered as ``pseudo'' pairs of training data. With the Cycle Consistency, the loss function of CycleGAN is defined as:
\begin{equation}
\begin{split}
& \mathcal{L}(G_{X \rightarrow Y}, G_{Y \rightarrow X}, D_{X}, D_{Y}) = 
\mathcal{L}(G_{X \rightarrow Y}, D_{Y}) \\
& \qquad\quad + \mathcal{L}(G_{Y \rightarrow X}, D_{X}) + \lambda\mathcal{L}_c(G_{X \rightarrow Y}, G_{Y \rightarrow X})
\end{split}
\label{eq:cyclegan}
\end{equation}
where 
\begin{equation}
\begin{split}
\mathcal{L}_c(G_{X \rightarrow Y}, G_{Y \rightarrow X})  
& = ||G_{Y \rightarrow X}(G_{X \rightarrow Y}(x))-x||_1 \\
& + ||G_{X \rightarrow Y}(G_{Y \rightarrow X}(y))-y||_1
\end{split}
\label{eq:consistancy}
\end{equation}
is the Cycle Consistency Loss. In our implementation, we adopt the network architecture of CycleGAN to train our conditional CycleGAN with the technical contributions described in the next subsections.

\begin{figure}[t]
	\begin{center}
		\includegraphics[width=0.8\linewidth]{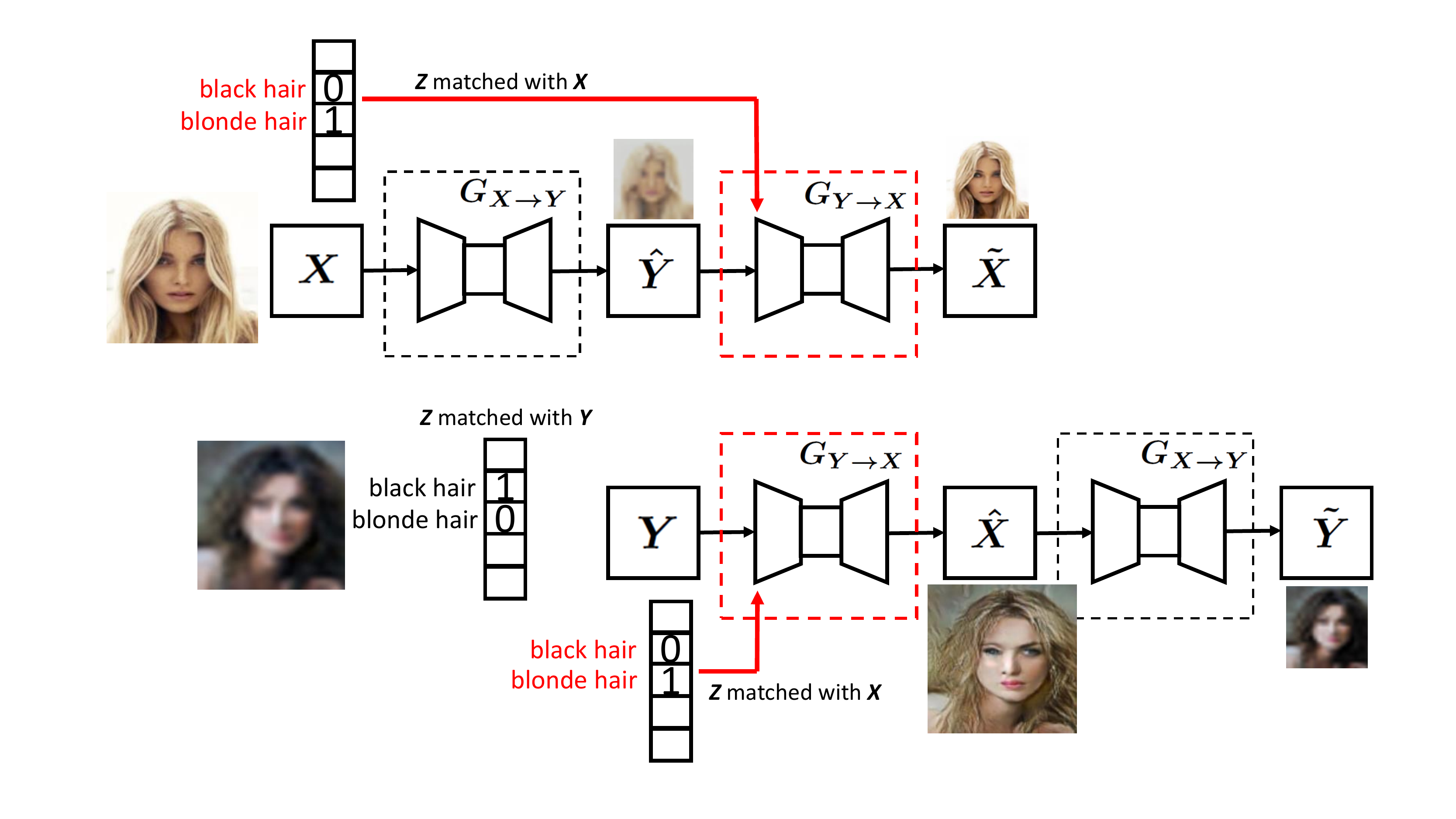} \\
	\end{center}
	\caption{Our Conditional CycleGAN for attribute-guided face generation. In contrast to the original CycleGAN, we embed an additional attribute vector $z$ (e.g., blonde hair) which is associated with the input attribute image $X$ to train a generator $G_{Y \to X}$ as well as the original $G_{X \to Y}$ to generate high-res face image $\hat{X}$ given the low-res input $Y$ and the attribute vector $z$. Note the discriminators $D_X$ and $D_Y$ are not shown for simplicity.}
	\label{fig:attributeGuided}
\end{figure}

\begin{figure}[t]
	\begin{center}
		\includegraphics[width=0.8\linewidth]{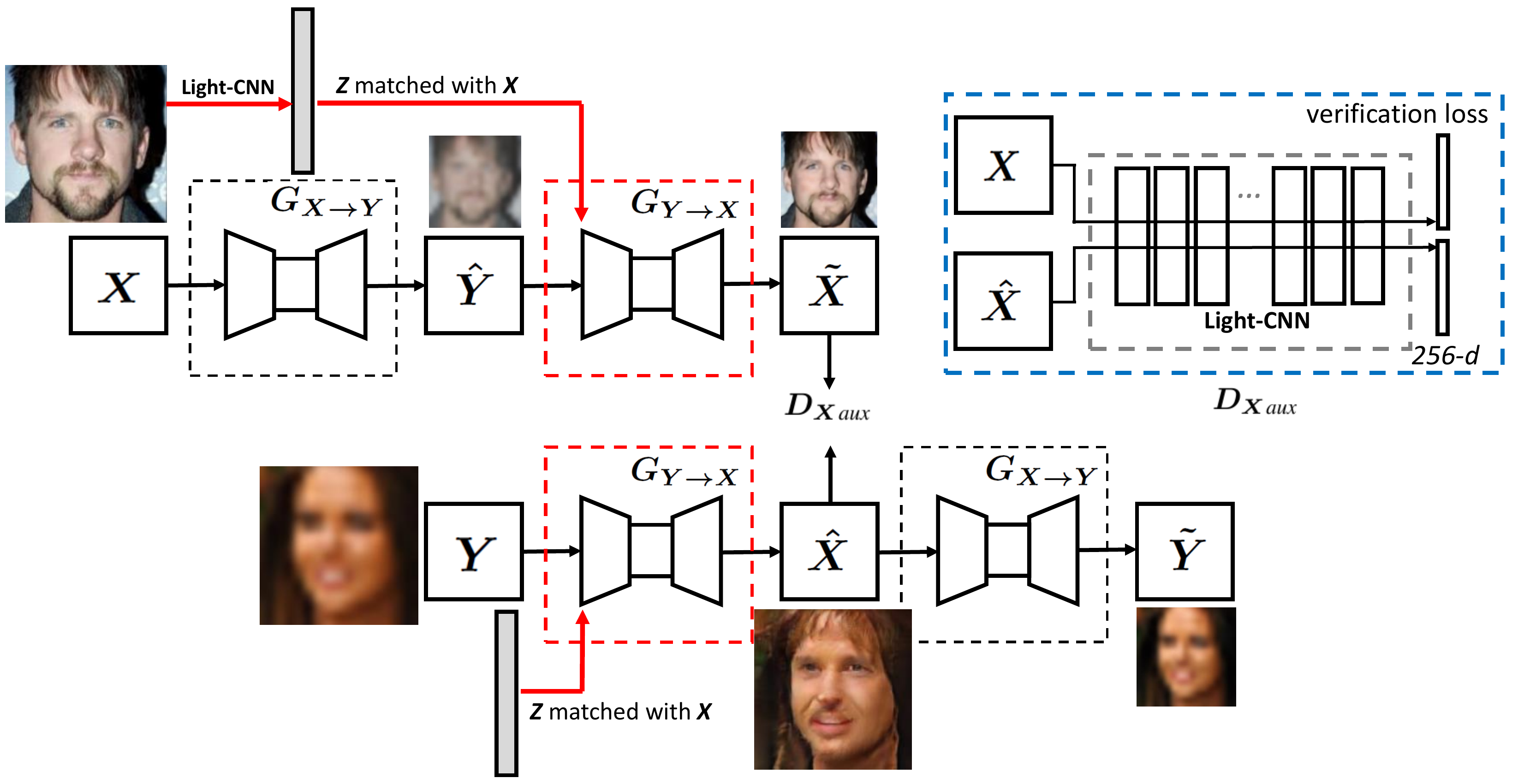} \\
	\end{center}
	\caption{Our Conditional CycleGAN for identity-guided face generation. Different from attribute-guided face generation, we incorporate a face verification network as both the source of conditional vector $z$ and the proposed identity loss in an auxiliary discriminator $D_{X_{\textit{aux}}}$. The network $D_{X_{\textit{aux}}}$ is pretrained. Note the discriminators $D_X$ and $D_Y$ are not shown for simplicity.}
	\label{fig:Idguided}
\end{figure}

\subsection{Attribute-guided Conditional CycleGAN}
\label{sec: attr}
We are interested in natural face image generation guided by user-supplied facial attributes to control
the high-res results.
To include conditional constraint into the CycleGAN network, the adversarial loss is modified to include the conditional feature vector $z$ as part of the input of the generator and intra layer to the discriminator as
\begin{equation}
\begin{split}
& \mathcal{L}(G_{(X,Z) \rightarrow Y}, D_{Y}) = 
\min_{\Theta_g} \max_{\Theta_d} \big\{ \mathbb{E}_{y,z} [\log D_{Y}(y,z)]  \\
& \qquad\qquad + \mathbb{E}_{x,z}[\log (1-D_{Y}(G_{(X,Z) \rightarrow Y}(x,z),z))\big\}
\end{split}
\label{eq:cgan}
\end{equation}
$\mathcal{L}(G_{(Y,Z) \rightarrow X}, D_{X})$ is defined analogously.

With the conditional adversarial loss, we modify the CycleGAN network as illustrated in Figure~\ref{fig:attributeGuided}. We follow~\cite{Perarnau2016_icgan} to pick 18 attributes as our conditional feature vector. Note that in our conditional CycleGAN, the attribute vector is associated with the input high-res face image (i.e., $X$), instead of the input low-res face image (i.e., $Y$). In each ``pair'' of training iteration, the same conditional feature vector is used to generate the high-res face image (i.e., $\hat{X}$). Hence, the generated intermediate high-res face image in the lower branch of Figure~\ref{fig:attributeGuided} will have different attributes from the corresponding ground truth high-res image. This is on purpose because the conditional discriminator network would enforce the generator network to utilize the information from the conditional feature vector. If the conditional feature vector always receives the correct attributes, the generator network would learn to skip the information in the conditional feature vector, since some of the attributes can be found in the low-res face image. 

\begin{algorithm}[t]
	\begin{algorithmic}[1]
		\footnotesize
		\caption{Conditional CycleGAN training procedure (using minibatch SGD as illustration)}
		\REQUIRE{Minibatch image sets $x \in X$ and $y \in Y$ in target and source domain respectively, attribute vectors $z$ matched with $x$ and mismatching $\hat{z}$, number of training batch iterations $S$} 
		\ENSURE{Update generator and discriminator weights $\theta_{g(X \to Y)}$, $\theta_{g(Y \to X)}$, $\theta_{d(X)}$, $\theta_{d(Y)}$}
		\STATE $\theta_{g(X \to Y)}$, $\theta_{g(Y \to X)}$, $\theta_{d(X)}$, $\theta_{d(Y)}$ $\leftarrow$ initialize network parameters 
		\FOR{$n=1$ to $S$}
		\STATE $\hat{y} \leftarrow G_{X \to Y}(x)$ \{Forward cycle $X \to Y$, fake $\hat{y}$\}
		\STATE $\tilde{x} \leftarrow G_{Y \to X}(\hat{y},z)$ \{Forward cycle $Y \to X$, reconstructed $\tilde{x}$\}
		\STATE $\hat{x} \leftarrow G_{Y \to X}(y,z)$ \{Backward cycle $Y \to X$, fake $\hat{x}$\}
		\STATE $\tilde{y} \leftarrow G_{X \to Y}(\hat{x})$ \{Backward cycle $X \to Y$, reconstructed $\tilde{y}$\}
		\STATE $\rho_r \leftarrow D_Y(y)$ \{Compute $D_Y$, real image\}
		\STATE $\rho_f \leftarrow D_Y(\hat{y})$ \{Compute $D_Y$, fake image\}
		\STATE $s_r \leftarrow D_X(y,z)$ \{Compute $D_X$, real image, right attribute\}
		\STATE $s_f \leftarrow D_X(\hat{y}, z)$ \{Compute $D_X$, fake image, right attribute\}
		\STATE $s_w \leftarrow D_X(y, \hat{z})$ \{Compute $D_X$, real image, wrong attribute\}
		\STATE $\mathcal{L}_{D_Y} \leftarrow \log(\rho_r) + \log(1-\rho_f)$ \{Compute $D_Y$ loss\}
		\STATE $\theta_{d(Y)} \leftarrow \theta_{d(Y)} - \alpha \nabla_{\theta_{d(Y)}} \mathcal{L}_{D_Y}$ \{Update on $D_Y$\}
		\STATE $\mathcal{L}_{D_X} \leftarrow \log(s_r) + \left[\log(1-s_f)+\log(1-s_w)\right]/2$ \\ \{Compute $D_X$ loss\}
		\STATE $\theta_{d(X)} \leftarrow \theta_{d(X)} - \alpha \nabla_{\theta_{d(X)}} \mathcal{L}_{D_X}$ \{Update on $D_X$\}
		\STATE $\mathcal{L}_c = \lambda_1 \|\tilde{x}-x\|_1 + \lambda_2 \|\tilde{y}-y\|_1$ \{Cycle consistency loss\}
		\STATE $\mathcal{L}_{G_{X \to Y}} \leftarrow \log(\rho_f)  + \mathcal{L}_c$ \{Compute $G_{X \to Y}$ loss\}
		\STATE $\theta_{g(X \to Y)} \leftarrow \theta_{g(X \to Y)} - \alpha \nabla_{\theta_{g(X \to Y)}} \mathcal{L}_{G_{X \to Y}}$ \\ \{Update on $G_{X \to Y}$\}
		\STATE $\mathcal{L}_{G_{Y \to X}} \leftarrow \log(s_f) + \mathcal{L}_c$ \{Compute $G_{Y \to X}$ loss\}
		\STATE $\theta_{g(Y \to X)} \leftarrow \theta_{g(Y \to X)} - \alpha \nabla_{\theta_{g(Y \to X)}} \mathcal{L}_{G_{Y \to X}}$ \\ \{Update on $G_{Y \to X}$\}
		\ENDFOR
	\end{algorithmic}
\end{algorithm}

In our implementation, the conditional feature vector is first replicated to match the size of the input image which is downsampled into a low-res. Hence, for $128 \times 128$ low-res input and $18$-dimensional feature vector, we have $18 \times 128 \times 128$ homogeneous feature maps after resizing. The resized feature is then concatenated with the {\em input} layer of the generator network to form a $(18+3) \times 128 \times 128$ tensor to propagate the inference of feature vector to the generated images. In the discriminator network, the resized feature (with size $18 \times 64 \times 64$) is concatenated with the {\em conv1} layer to form a $(18+64) \times 64 \times 64$ tensor. 

Algorithm~1 describes the the whole training procedure, with the network illustrated 
in Figure~\ref{fig:attributeGuided}.
In order to train the conditional GAN network, only the correct pair of groundtruth high-res face image and the associated attribute feature vector are treated as positive examples. The generated high-res face image with the associated attribute feature vector, and the groundtruth high-res face image with randomly sampled attribute feature vector are both treated as negative examples. In contrast to traditional CycleGAN, we use conditional adversarial loss and conditional cycle consistency loss for updating the networks.

\subsection{Identity-guided Conditional CycleGAN} 
\label{sec: id}
To demonstrate the efficacy of our conditional CycleGAN guided by control attributes, we specialize it into identity-guided face image generation. We utilize the feature vector from a face verification network, i.e. Light-CNN~\cite{DBLP:journals/corr/WuHS15_lightcnn} as the conditional feature vector. The identity feature vector is a 256-D vector from the ``Light CNN-9 model". Compared with another state-of-the-art 
FaceNet~\cite{Schroff_2015_CVPR_19}, which returns a 1792-D face feature vector for each face image, 
the 256-D representation of light-CNN obtains state-of-the-art results while it has fewer parameters 
and runs faster. Though among the best single models, the Light-CNN can be easily replaced by other face verification networks like FaceNet or VGG-Face.

\subsubsection{Auxiliary Discriminator}
In our initial implementation, we follow the same architecture and training strategy to train the conditional CycleGAN for identity-guided face generation. However, we found that the trained network does not generate good results (sample shown in Figure~\ref{fig:component} (d)). We believe this is because the discriminator network is trained from scratch, and the trained discriminator network is not as powerful as the light-CNN which was trained from million pairs of face images. 

Thus, we add an auxiliary discriminator $D_{X_{\textit{aux}}}$ on top of the conditional generator $G_{Y \to X}$ in parallel with the discriminator network $D_X$ so there are two discriminators for $G_{Y \to X}$, while the discriminator for $G_{X \to Y}$ remains the same (as illustrated in Figure~\ref{fig:Idguided}). Our auxiliary discriminator takes an input of the generated high-res image $\hat{X}$ or the ground truth image $X$, and outputs a feature embedding. We reuse the pretrained Light-CNN model for our auxiliary discriminator, the activation of the second last layer: the 256-D vector same as our conditional vector $Z$. 

Based on the output of the auxiliary discriminator, we define an identity loss to better guide the learning of the generator. Here we use the L1 loss of the output 256-D vectors as our identity loss. The verification errors from the auxiliary discriminator is back propagated concurrently with the errors from the discriminator network. With the face verification loss, we are able to generate high quality high-res face images matching the identity given by the conditional feature vector. 
As shown in the running example in Figure~\ref{fig:Idguided}, the lady's face is changed to a man's face whose
identify is given by the light-CNN feature. 

%-------------------------------------------------------------------------

\section{Experiments}

We use two image datasets, MNIST (for sanity check) and CelebA~\cite{CelebA} (for face image generation) to evaluate our method. The MNIST is a digit dataset of 60,000 training and 10,000 testing images. Each image is a 28$\times$28 black and white digit image with the class label from 0 to 9. The CelebA is a face dataset of 202,599 face images, with 40 different attribute labels where each label is a binary value. We use the aligned and cropped version, with 182K images for training and 20K for testing. To generate low-res images, we downsampled the images in both datasets by a factor of 8, and we separate the images such that the high-res and low-res training images are non-overlapping.

%\begin{figure*}[t]
%	\begin{center}
%		\includegraphics[width=1.0\linewidth]{figures/icgan.pdf} \\
%	\end{center}
%	\caption{Comparison with IcGAN~\cite{icgan} on all 18 attributes. Top: Our results. Bottom: Results by IcGAN. Column 1 are the ground truth images and column 2 the generated images. Column %3 to 20 are results when flipping one of the 18 attributes.}
%	\label{fig:icgan}
%\end{figure*}

\subsection{MNIST}
\begin{figure}[t]
	\begin{center}
		\includegraphics[width=0.6\linewidth]{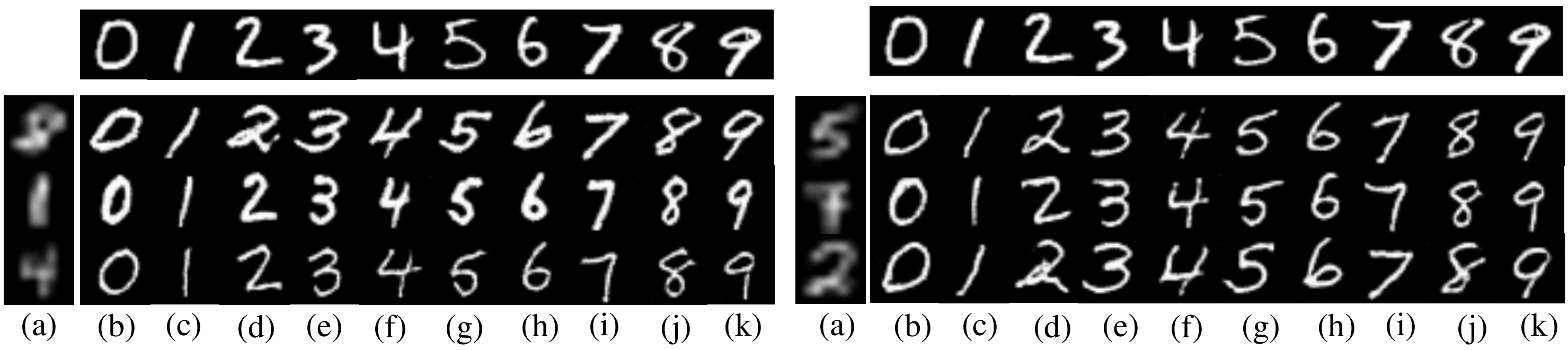} \\
	\end{center}
	\caption{From the low-res digit images (a), we can generate high-res digit images (b) to (k) subject to the conditional constrain from the digit class label in the first row.}
	\label{fig:digit}
\end{figure}

We first evaluate the performance of our method on MNIST dataset.  The
conditional feature vector is
the class label of digits. As shown in Figure~\ref{fig:digit}, our method can generate
high-res digit images
from the low-res inputs.  Note that the generated high-res
digit follows the
given class label when there is conflict between the low-res image
and feature vector.
This is desirable, since the conditional constraint consumes large weights
during the training.
This sanity check also verifies that we can impose conditional constraint
into the CycleGAN
network.

In addition to the label changes based on the high-res identity inputs, we observe that 
the generated high-res images inherit the appearance in the low-res inputs 
such as the orientation and thickness. For the `8' example in Figure~\ref{fig:digit} the 
outputs share the same slanting orientation with the low-res `8' which is tilted to the right. 
In the next row the outputs adopt the thickness of the input, that is, the relatively thick 
stroke presented by the low-res `1'. This is a good indicator of the ability of our trained generator: 
freedom in changing labels based on the high-res images presented as identity attribute, 
while preserving the essential appearance feature presented by the low-res inputs. 

\begin{figure}[t]
	\begin{center}
		\includegraphics[width=0.7\linewidth]{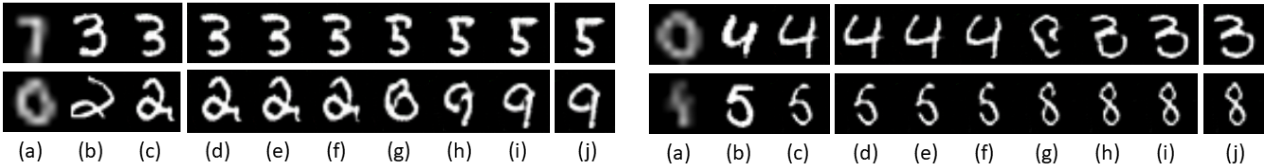} \\
	\end{center}
	\caption{Interpolation results of digits. Given the low-res inputs in (a), we randomly sample two digits (b) and (j). (c) is the generated results from (a) conditioned on the attribute of (b). Corresponding results of interpolating between attributes of (b) and (j) are shown in (d) to (i). We interpolate between the binary vectors of the digits.}
	\label{fig:interpolate_digit}
\end{figure}

\begin{figure}[!h]
	\begin{center}
		\includegraphics[width=1.0\linewidth]{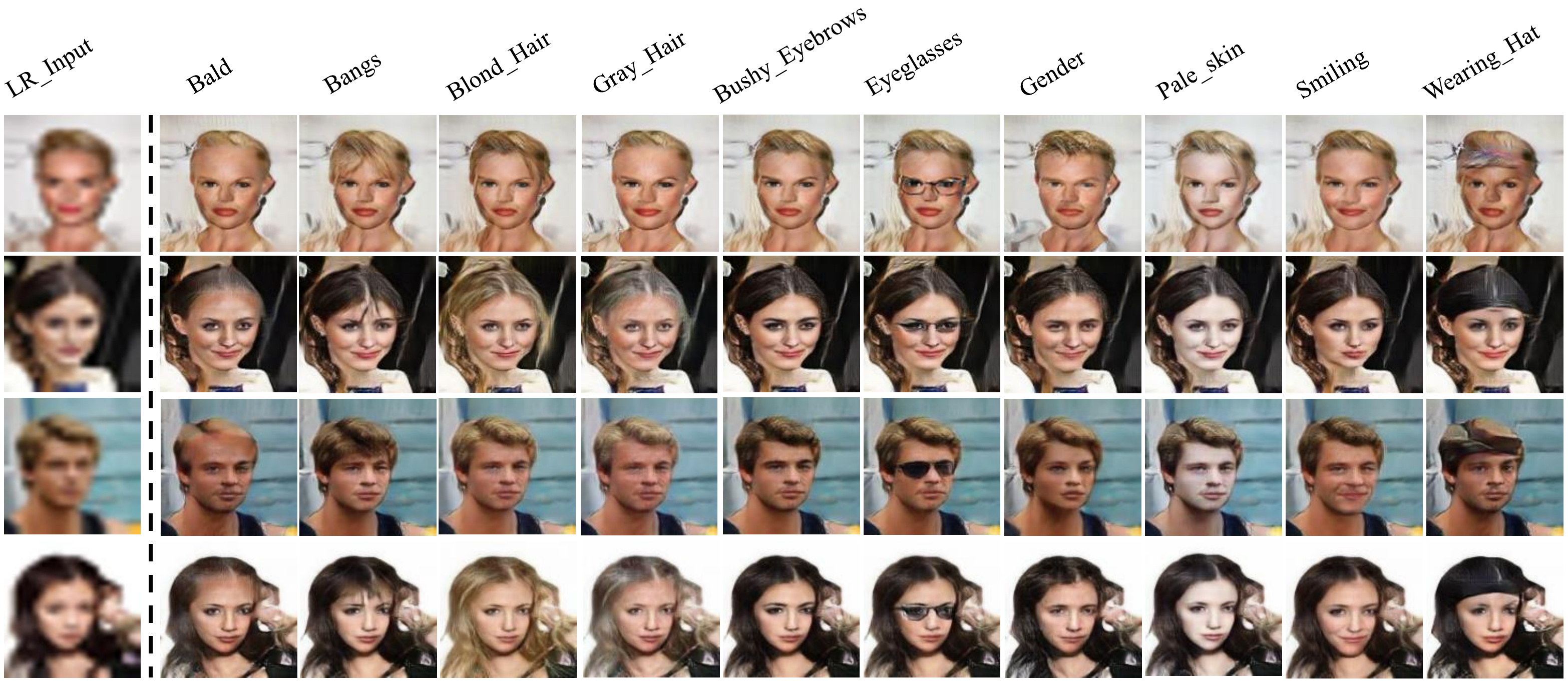} \\
	\end{center}
	\caption{Attribute-guided face generation. We flip one attribute label for each generated high-res face images, given the low-res face inputs. The 10 labels are: Bald, Bangs, Blond$\_$Hair, Gray$\_$Hair, Bushy$\_$Eyebrows, Eyeglasses, Male, Pale$\_$Skin, Smiling, Wearing$\_$Hat.}
	\label{fig:attribute}
\end{figure}

Apart from generating high-res digit images from the low-res inputs, we also perform linear interpolation between two high-res images (as identity features) to show our model is able to learn the digit representation. Specifically, we interpolate between the respective binary vectors of the two digits. Sample results are shown in Figure~\ref{fig:interpolate_digit}.

% \begin{figure}[t]
% 	\begin{center}
% 		\includegraphics[width=1.0\linewidth]{figures/smile.pdf} \\
% 	\end{center}
% 	\caption{Comparison results of \textit{Smiling} attribute-guided generation with~\cite{kim2017unsupervised}. Four source images are shown in top row. Images with green, blue and red bounding boxes indicates reference attribute images, transferred results by~\cite{kim2017unsupervised} and results by our method, respectively.}
% 	\label{fig:smile}
% \end{figure}

% \begin{figure}[!h]
% 	\begin{center}
% 		\includegraphics[width=1.0\linewidth]{figures/hair.pdf} \\
% 	\end{center}
% 	\caption{Comparison results of \textit{Hair Color} attribute-guided generation with~\cite{kim2017unsupervised}. Four source images are shown in top row. Images with green, blue and red bounding boxes indicates reference attribute images, transferred results by~\cite{kim2017unsupervised} and results by our method, respectively.}
% 	\label{fig:hair}
% \end{figure}

\begin{figure}[!h]
	\begin{center}
		\includegraphics[width=1.0\linewidth]{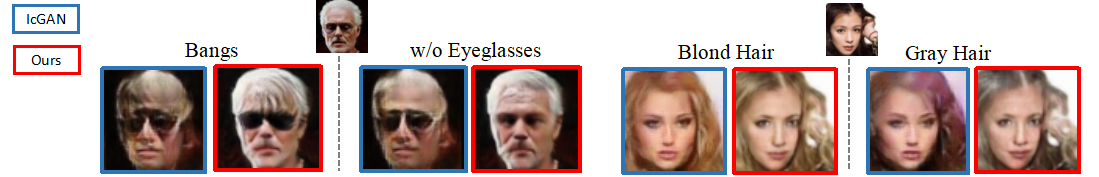} \\
	\end{center}
	\caption{Comparison with~\cite{Perarnau2016_icgan} by swapping facial attributes. Four paired examples are shown. Generally, our method can generate much better images compared to~\cite{Perarnau2016_icgan}.}
	\label{fig:icgan}
\end{figure}

\begin{figure}[htb]
	\begin{center}
		\includegraphics[width=0.9\linewidth]{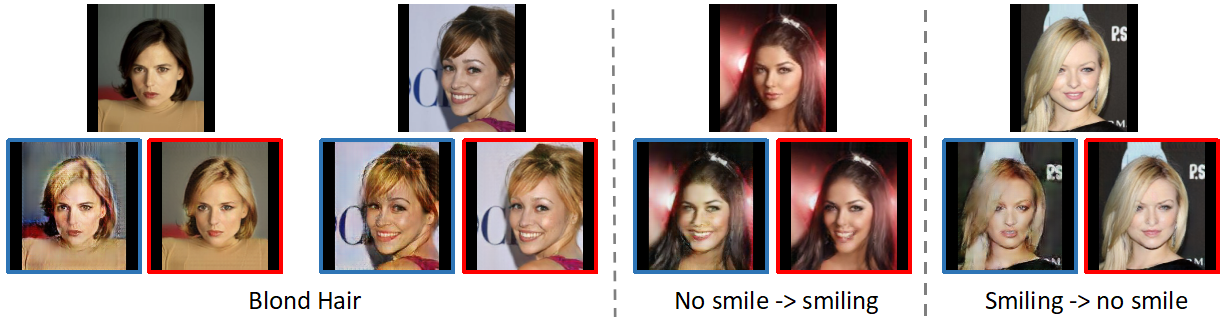} \\
	\end{center}
	\caption{Comparison results with~\cite{kim2017learning}. Four source images are shown in top row. Images with blue and red bounding boxes indicates transferred results by~\cite{kim2017learning} and results by our method, respectively.}
	\label{fig:disco}
\end{figure}

\begin{table}[t] 
	\centering
	\caption{SSIM on CelebA test sets.}
	\begin{tabular}{c||c|c|c}
		\hline
		Method &   Conditional GAN~\cite{Perarnau2016_icgan} & Unsupervised GAN~\cite{kim2017learning} & Conditional CycleGAN \\
		\hline
		SSIM & 0.74 & 0.87 & \textbf{0.92} \\
		\hline
	\end{tabular}
	\label{tab:ssim}
\end{table}

\begin{figure}[t]
	\begin{center}
		\includegraphics[width=1.0\linewidth]{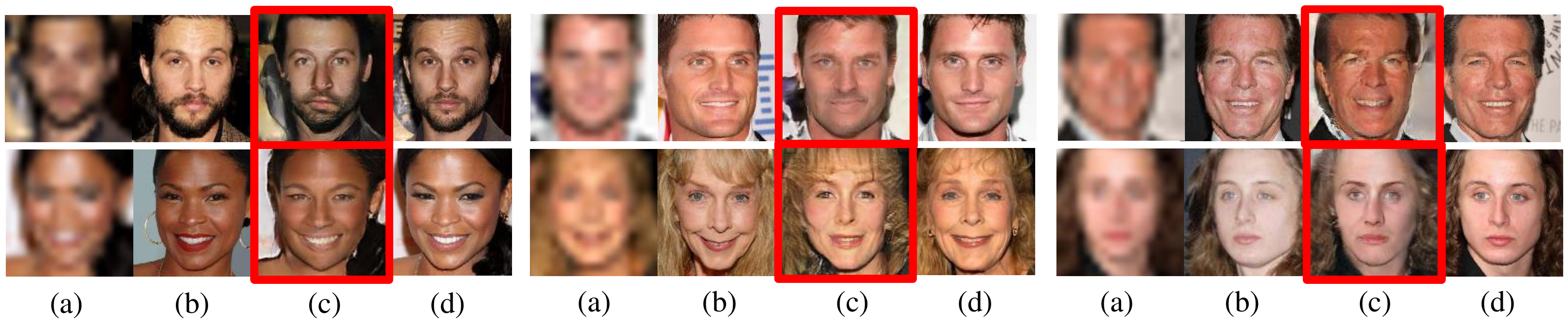} \\
	\end{center}
	\caption{Identity-guided face generation results on low-res input and high-res identity of the same person, i.e., identity-preserving face superresolution. (a) low-res inputs; (b) input identity of the same person; (c) our high-res face outputs (red boxes) from (a); (d) the high-res ground truth of (a).}
	\label{fig:IdSR_same}
\end{figure}

\begin{figure}[!h]
	\begin{center}
		\includegraphics[width=1.0\linewidth]{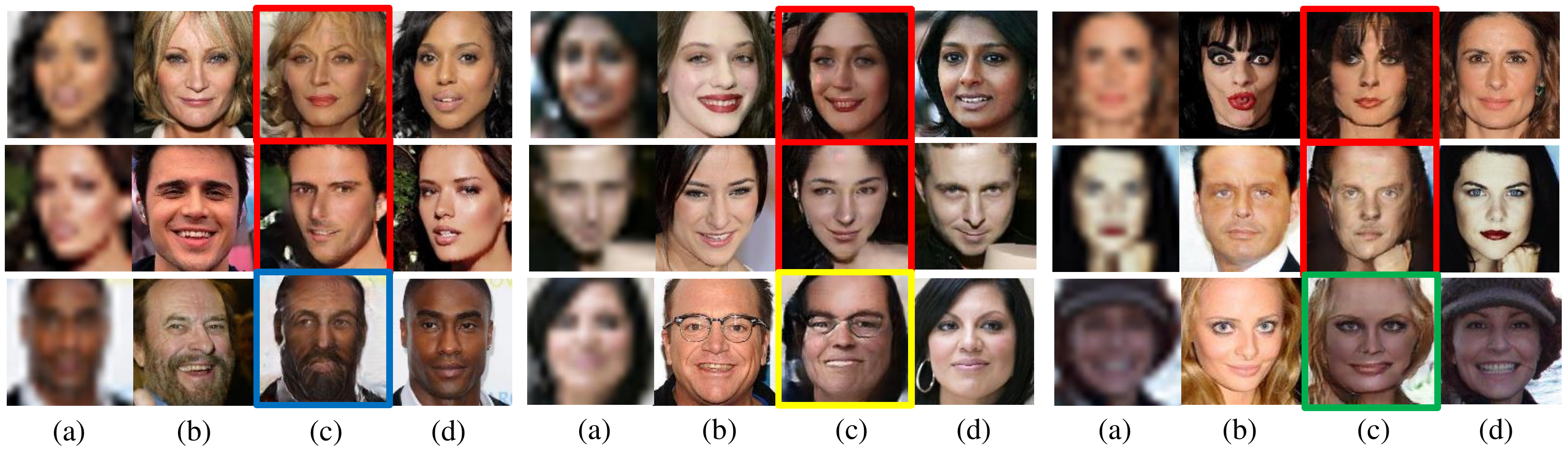} \\
	\end{center}
	\caption{Identity-guided face generation results on different persons. The last row shows some challenging examples, e.g., , the occluded forehead in low-res input is recovered (example in the green box). (a) low-res inputs provide overall shape constraint; (b) identity to be transferred; (c) our high-res face outputs (red boxes) from (a) where the man/woman's identity in (b) is transferred; (d) the high-res ground truth of (a).}
	\label{fig:IdSR_diff}
\end{figure}

\subsection{Attribute-guided Face Generation}
Figure~\ref{fig:attribute} shows sample results for attribute guided face
generation.
Recall the condition is encoded as a 18-D vector.  The 10 results shown in
the figure are generated with one attribute label flipped in their corresponding
condition vector in conditional CycleGAN.  Our generated results conditioned on
attributes such as {\sc Bangs}, {\sc Blond$\_$Hair}, {\sc Bushy$\_$Eyebrows}, {\sc Gender}, {\sc Pale$\_$Skin} are quite convincing.\\

\noindent{\bf Comparison with Conditional GAN.}
We first compare with conditional GAN framework~\cite{Perarnau2016_icgan} under the attribute-guided face generation framework. Visualizations are shown in Figure~\ref{fig:icgan}. Generally, our method can generate much better images compared to the competitor, e.g., our methods successfully removes the eyeglasses, as well as generates the right color of hairs. Note that~\cite{Perarnau2016_icgan} generates different persons while ours are faithful to the inputs.\\

\noindent{\bf Comparison with Unsupervised GAN.}
We further compare with~\cite{kim2017learning}, which is also a unpaired image-to-image translation method. Comparison results are shown in Figure~\ref{fig:disco}. Note we only provide part of the attribute results according to their paper for fair comparison.\\

\noindent{\bf Quantitative Comparison.}
To quantitatively evaluate the generated results, we use structural similarity
(SSIM)~\cite{wang2004image}, which is a widely used image quality metric that correlates well with human visual perception. SSIM ranges from 0 to 1, higher is better. The SSIM of our method, as well as conditional GAN~\cite{Perarnau2016_icgan} and unsupervised GAN~\cite{kim2017learning}, is shown in Table~\ref{tab:ssim} for the generated images. 

Our method outperforms~\cite{Perarnau2016_icgan} and~\cite{kim2017learning} in two aspects: (i) In the unsupervised GAN setting, compared with~\cite{kim2017learning}, our method shows significant performance gain with the proposed attributed guided framework. (ii) compared with conditional GAN~\cite{Perarnau2016_icgan}, the performance gain is even larger with the help of the cyclic network architecture of our conditional CycleGAN. 

\subsection{Identity-guided Face Generation}
Figure~\ref{fig:IdSR_same} and Figure~\ref{fig:IdSR_diff} show sample face generation results where the identity face features are
respectively from the same and {\em different} persons.  There are two interesting points to note:  First,
the generated high-res results (c) bear high resemblance to the target identity images (b)
from which the identity features are computed using
Light-CNN. The unique identity features transfer well from (b) to (c), e.g., challenging gender change in the second row of Figure~\ref{fig:IdSR_diff}. In the last row, the facial attributes, e.g. beard (example in the blue box), eyeglasses (example in the yellow box) are considered as parts of the identity and faithfully preserved by our model in the high-res outputs, even though the low-res inputs do not have such attributes. The occluded forehead in low-res input (example in the green box) is recovered. 
Second, the low-res inputs provide overall shape constraint. The head pose and facial expression of the generated high-res
images (c) adopt those in the {\em low}-res inputs (a). Specifically,
refer to the example inside the blue box in the last row of Figure~\ref{fig:IdSR_diff}, where 
(b) shows target identity, i.e. man smiling while low-res input (a) shows another man with closed mouth.
The generated high-res image in (c) preserves the identity in (b)
while the pose of the head follows the input and the mouth is closed as well. 
%Head pose and facial expression are
%{\em not} parts of the identity here, and the original facial expression in the
%low-res input, that is, closed mouth, should be
%preserved in the high-res output.

\begin{figure}[t]
	\begin{center}
		\includegraphics[width=1.0\linewidth]{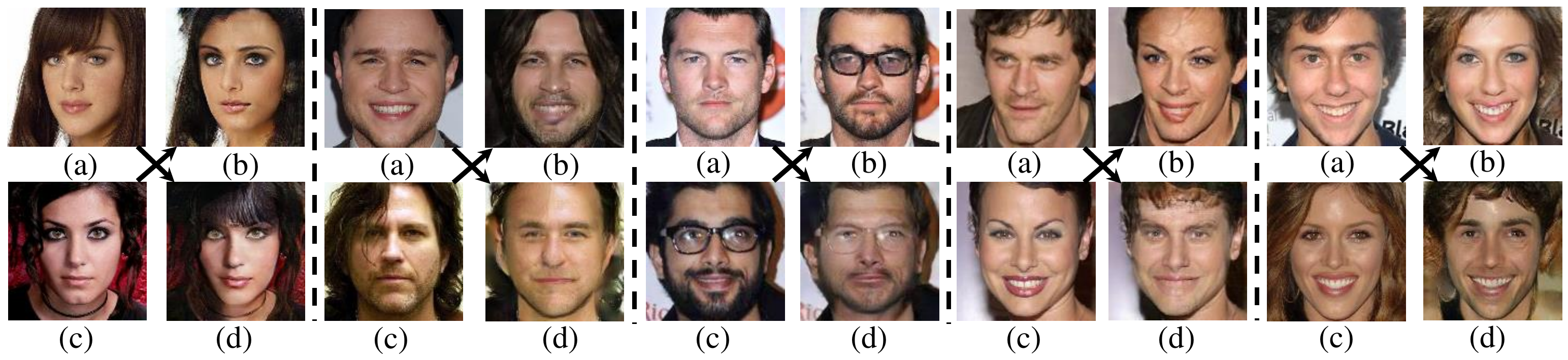} \\
	\end{center}
	\caption{Face swapping results within the high-res domain. (a)(c) are inputs of two different persons; (b)(d) their face swapping results. The black arrows indicate the guidance of identity, i.e. (d) is transformed from (c) under the identity constraint of (a). Similarly, (b) is transformed from (a) under the identity of (c). Note how our method transforms the identity by altering the appearance of eyes, eyebrows, hairs etc, while keeping other factors intact, e.g., head pose, shape of face and facial expression.}
	\label{fig:a2a}
\end{figure}

\subsection{Face Swapping within the High-Res Domain}
We demonstrate an interesting application {\em face swapping} where {\em
	both} the input and identity images are high-res images.  Here, we want to
swap the identity while preserving {\em all} facial details including
subtle crease lines and expression, thus both the identity image and the
input image must be high-res images. We adopt our identity-guided conditional CycleGAN and utilize Light-CNN as both the source of the identity features and face verification loss. Our face swapping results are shown in Figure~\ref{fig:a2a}. As illustrated, our method swaps the identity by transferring the appearance of eyes, eyebrows, hairs, etc, while keeping other factors intact, e.g., head pose, shape of face and facial expression. 
Without multiple steps (e.g., facial landmark detection followed by warping and blending) in traditional techniques, our identity-guided conditional CycleGAN can still achieve high levels of photorealism of the face-swapped images.

%For further study we provide qualitative comparison results between our two applications, face swapping and identity-preserving face generation, with the same target identity image. Results are shown in Figure~\ref{fig:compare}. We can learn from the figures that there are more constraints on face swapping results, e.g., in the first row (e), the eyes and eyebrows resemble high res in (c), while in (d) it generates blonde hair and different eyebrows from (b). 

%\begin{figure}[t]
%	\begin{center}
%		\includegraphics[width=0.55\linewidth]{figures/compare.pdf} \\
%	\end{center}
%	\caption{Comparison results between SR and face swapping. (a) Identity images. (b) Low-res input for SR. (c) Input in high-res domain for face swapping. (d) Generated high-res results from (b). (e) Face swapping results from (c).}
%	\label{fig:compare}
%\end{figure}

\begin{figure}[t]
	\begin{center}
		\includegraphics[width=0.45\linewidth]{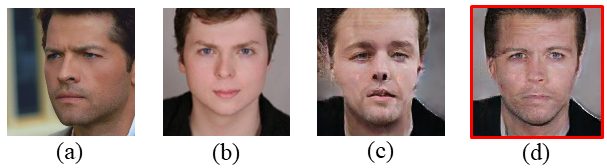} \\
	\end{center}
	\caption{Results without (c) and with (d) face verification loss. (a) is target identity image to be transferred and (b) is input image. The loss encourages subtle yet important improvement in photorealism, e.g. the eyebrows and eyes in (c) resemble the target identity in (a) by adding the face verification loss.}
	\label{fig:component}
\end{figure}

Figure~\ref{fig:component} compares face swapping results of our models trained with and without the face verification loss in the auxiliary discriminator. The difference is easy to recognize, and adding face verification loss has a perceptual effect of improving the photorealism of swapped-face image. In this example, the eyebrows and eyes are successfully transformed to the target identity with the face verification loss.

\subsection{Frontal Face Generation}

\begin{figure}[t]
	\begin{center}
		\includegraphics[width=1.0\linewidth]{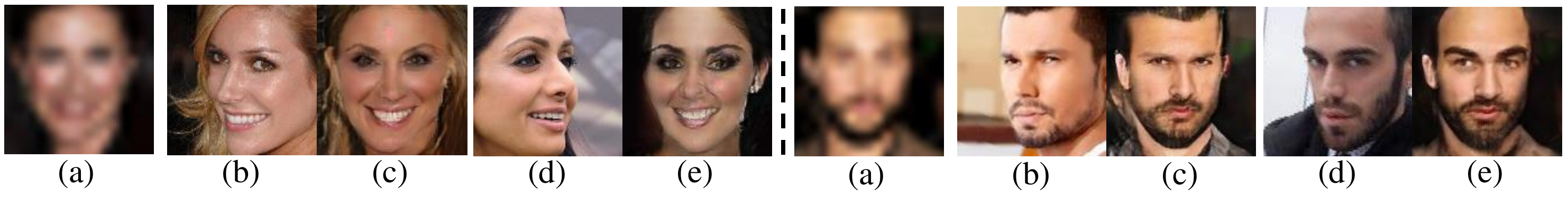} \\
	\end{center}
	\caption{Frontal face generation. Given a low-res template (a), our method can generate corresponding frontal faces from different side faces, e.g., (b) to (c), (d) to (e). }
	\label{fig:frontal}
\end{figure}

Another application of our model consists of generating images of frontal faces from face images in other orientations. By simply providing a low-res frontal face image and adopting our identity-guided conditional CycleGAN model, we can generate the corresponding high-res frontal face images given 
side-face images as high-res face attributes. Figure~\ref{fig:frontal} shows sample results on our
frontal face image generation. Note that our frontal face generation is end-to-end and free of human intervention, thus setting it apart from related works of frontalizing the face by landmark detection, warping and blending etc. given a frontal pose.

\subsection{Interpolating Conditional Vector}

\begin{figure}[t]
	\begin{center}
		\includegraphics[width=1.0\linewidth]{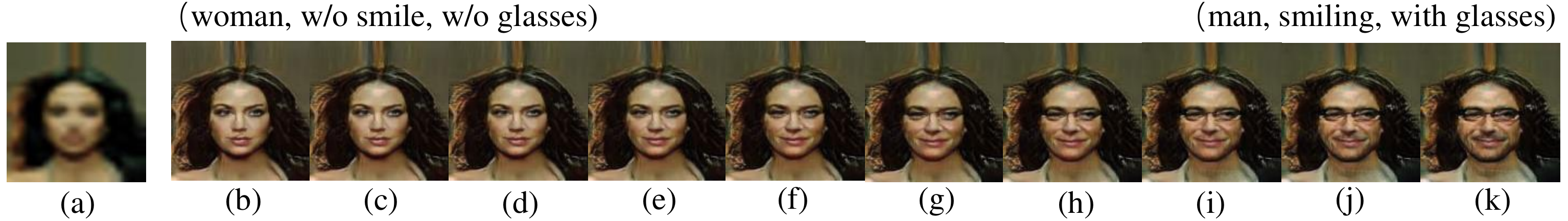} \\
	\end{center}
	\caption{Interpolation results of the attribute vectors. (a) Low-res face input; (b) generated high-res face images; (c) to (k) interpolated results. Attributes of source and destination are shown in text.}
	\label{fig:interpolate}
\end{figure}

\begin{figure}[!h]
	\begin{center}
		\includegraphics[width=1.0\linewidth]{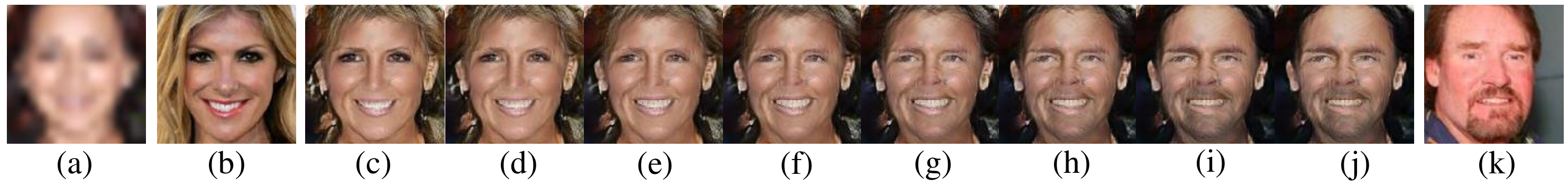} \\
	\end{center}
	\caption{Interpolating results of the identity feature vectors. Given the low-res input in (a), we randomly sample two target identity face images (b) and (k). (c) is the generated face from (a) conditioned on the identity in (b) and (d) to (j) are interpolations. }
	\label{fig:interpolate_id}
\end{figure}

We further explore the conditional attribute vector by linearly interpolating between two different attribute vectors. Figure~\ref{fig:interpolate} shows that all the interpolated faces are visually plausible with smooth transition among them, which is a convincing demonstration that the model generalizes well the face representation instead of just directly memorizes the training samples.

Similar to interpolating the attribute vectors, we experiment with interpolating the 256-D identity feature vectors under our identity-guided conditional model. We randomly sample two high-res face images and interpolate between the two identity features. Figure~\ref{fig:interpolate_id} indicates that our model generalizes properly the face representation given the conditional feature vectors.

\section{Conclusion}

We have presented the Conditional CycleGAN for attribute-guided and
identity-guided face image generation. Our technical contribution 
consists of the conditional CycleGAN to guide the face image generation 
process via easy user input of complex attributes for generating high 
quality results.  In the attribute-guided conditional
CycleGAN, the adversarial loss is modified to include a conditional
feature vector as parts of the inputs to the generator and discriminator
networks.  We utilize the feature vector from light-CNN in
identity-guided conditional CycleGAN.  We have presented the first
but significant results on identity-guided and attribute-guided face 
image generation. In the future, we will
explore how to further improve the results and extend the work
to face video  generation. 

\subsubsection*{Acknowledgement}
This work was supported in part by Tencent Youtu.


\bibliographystyle{splncs04}
\bibliography{faceSR}

\begin{thebibliography}{10}
\providecommand{\url}[1]{\texttt{#1}}
\providecommand{\urlprefix}{URL }
\providecommand{\doi}[1]{https://doi.org/#1}

\bibitem{Choi_2018_CVPR}
Choi, Y., Choi, M., Kim, M., Ha, J.W., Kim, S., Choo, J.: Stargan: Unified
  generative adversarial networks for multi-domain image-to-image translation.
  In: The IEEE Conference on Computer Vision and Pattern Recognition (CVPR)
  (June 2018)

\bibitem{Dong2014_6}
Dong, C., Loy, C.C., He, K., Tang, X.: Learning a Deep Convolutional Network
  for Image Super-Resolution (2014)

\bibitem{DBLP:journals/corr/DongLT16}
Dong, C., Loy, C.C., Tang, X.: Accelerating the super-resolution convolutional
  neural network. CoRR  \textbf{abs/1608.00367} (2016),
  \url{http://arxiv.org/abs/1608.00367}

\bibitem{GAN}
Goodfellow, I., Pouget-Abadie, J., Mirza, M., Xu, B., Warde-Farley, D., Ozair,
  S., Courville, A., Bengio, Y.: Generative adversarial nets. In: NIPS, pp.
  2672--2680 (2014),
  \url{http://papers.nips.cc/paper/5423-generative-adversarial-nets.pdf}

\bibitem{2017arXiv170404086H}
{Huang}, R., {Zhang}, S., {Li}, T., {He}, R.: {Beyond Face Rotation: Global and
  Local Perception GAN for Photorealistic and Identity Preserving Frontal View
  Synthesis}. ArXiv e-prints  (Apr 2017)

\bibitem{DBLP:journals/corr/IsolaZZE16}
Isola, P., Zhu, J., Zhou, T., Efros, A.A.: Image-to-image translation with
  conditional adversarial networks. CoRR  \textbf{abs/1611.07004} (2016),
  \url{http://arxiv.org/abs/1611.07004}

\bibitem{DBLP:journals/corr/JohnsonAL16}
Johnson, J., Alahi, A., Li, F.: Perceptual losses for real-time style transfer
  and super-resolution. CoRR  \textbf{abs/1603.08155} (2016),
  \url{http://arxiv.org/abs/1603.08155}

\bibitem{DBLP:conf/cvpr/KimLL16}
Kim, J., Lee, J.K., Lee, K.M.: Deeply-recursive convolutional network for image
  super-resolution. In: CVPR. pp. 1637--1645 (2016)

\bibitem{kim2017learning}
Kim, T., Cha, M., Kim, H., Lee, J., Kim, J.: Learning to discover cross-domain
  relations with generative adversarial networks. ICML  (2017)

\bibitem{DBLP:journals/corr/LedigTHCATTWS16_7}
Ledig, C., Theis, L., Huszar, F., Caballero, J., Aitken, A.P., Tejani, A.,
  Totz, J., Wang, Z., Shi, W.: Photo-realistic single image super-resolution
  using a generative adversarial network. CoRR  \textbf{abs/1609.04802} (2016),
  \url{http://arxiv.org/abs/1609.04802}

\bibitem{2017arXiv170501088L}
{Liao}, J., {Yao}, Y., {Yuan}, L., {Hua}, G., {Kang}, S.B.: {Visual Attribute
  Transfer through Deep Image Analogy}. SIGGRAPH  (2017)

\bibitem{CelebA}
Liu, Z., Luo, P., Wang, X., Tang, X.: Deep learning face attributes in the
  wild. CoRR  \textbf{abs/1411.7766} (2014)

\bibitem{Perarnau2016_icgan}
Perarnau, G., van~de Weijer, J., Raducanu, B., \'Alvarez, J.M.: {Invertible
  Conditional GANs for image editing}. In: NIPS Workshop on Adversarial
  Training (2016)

\bibitem{DBLP:journals/corr/RadfordMC15}
Radford, A., Metz, L., Chintala, S.: Unsupervised representation learning with
  deep convolutional generative adversarial networks. CoRR
  \textbf{abs/1511.06434} (2015), \url{http://arxiv.org/abs/1511.06434}

\bibitem{Schroff_2015_CVPR_19}
Schroff, F., Kalenichenko, D., Philbin, J.: Facenet: A unified embedding for
  face recognition and clustering. In: CVPR (June 2015)

\bibitem{espcn}
Shi, W., Caballero, J., Huszar, F., Totz, J., Aitken, A.P., Bishop, R.,
  Rueckert, D., Wang, Z.: Real-time single image and video super-resolution
  using an efficient sub-pixel convolutional neural network. In: CVPR. pp.
  1874--1883 (2016)

\bibitem{DBLP:journals/corr/WangWWL16}
Wang, Y., Wang, L., Wang, H., Li, P.: End-to-end image super-resolution via
  deep and shallow convolutional networks. CoRR  \textbf{abs/1607.07680}
  (2016), \url{http://arxiv.org/abs/1607.07680}

\bibitem{wang2004image}
Wang, Z., Bovik, A.C., Sheikh, H.R., Simoncelli, E.P.: Image quality
  assessment: from error visibility to structural similarity. IEEE transactions
  on image processing  \textbf{13}(4),  600--612 (2004)

\bibitem{DBLP:journals/corr/WuHS15_lightcnn}
Wu, X., He, R., Sun, Z.: A lightened {CNN} for deep face representation. CoRR
  \textbf{abs/1511.02683} (2015), \url{http://arxiv.org/abs/1511.02683}

\bibitem{yi2017dualgan}
Yi, Z., Zhang, H., Gong, P.T., et~al.: Dualgan: Unsupervised dual learning for
  image-to-image translation. arXiv preprint arXiv:1704.02510  (2017)

\bibitem{zhang2017stackgan}
Zhang, H., Xu, T., Li, H., Zhang, S., Huang, X., Wang, X., Metaxas, D.:
  Stackgan: Text to photo-realistic image synthesis with stacked generative
  adversarial networks. In: IEEE Int. Conf. Comput. Vision (ICCV). pp.
  5907--5915 (2017)

\bibitem{multiview}
{Zhao}, B., {Wu}, X., {Cheng}, Z.Q., {Liu}, H., {Feng}, J.: {Multi-View Image
  Generation from a Single-View}. ArXiv e-prints  (Apr 2017)

\bibitem{CycleGAN2017}
Zhu, J.Y., Park, T., Isola, P., Efros, A.A.: Unpaired image-to-image
  translation using cycle-consistent adversarial networks. In: Computer Vision
  (ICCV), 2017 IEEE International Conference on (2017)

\end{thebibliography}

\end{document}